\documentclass{article}
\usepackage[utf8]{inputenc}
\usepackage{booktabs}
\usepackage{glossaries}
\usepackage{cite}
\usepackage[ruled]{algorithm2e}
\usepackage{amssymb}
\usepackage{url}
\usepackage{graphicx}
\usepackage{caption}
\usepackage{subcaption}
\usepackage{authblk}

\newacronym{mri}{MRI}{Magnetic Resonance Imaging}
\newacronym{dc}{DC}{Dice Coefficient}
\newacronym{crf}{CRF}{Conditional Random Field}
\newacronym{brats_2015}{BraTS 2015}{Multimodal Brain Tumor Segmentation Challenge 2015}
\newacronym{hgg}{HGG}{High-Grade Glioma}
\newacronym{rnn}{RNN}{Recurrent Neural Network}
\newacronym{cnn}{CNN}{Convolutional Neural Network}
\newacronym{map}{MAP}{maximum a posteriori}
\newacronym{fcnn}{FCNN}{Fully-Convolutional Neural Network}

\title{Conditional Random Fields as Recurrent Neural Networks for 3D Medical Imaging Segmentation}

\author[1]{Miguel Monteiro\thanks{Corresponding author: \texttt{mab.mtr@gmail.com} }}
\author[2,3]{Mário A. T. Figueiredo}
\author[1,2]{Arlindo~L.~Oliveira}

\affil[1]{INESC-ID}
\affil[2]{Instituto Superior Técnico, Universidade de Lisboa}
\affil[3]{Instituto de Telecomunicações}

\date{}

\begin{document}

\maketitle

\begin{abstract}
    \noindent The Conditional Random Field as a Recurrent Neural Network layer is a recently proposed algorithm meant to be placed on top of an existing Fully-Convolutional Neural Network to improve the quality of semantic segmentation.
    In this paper, we test whether this algorithm, which was shown to improve semantic segmentation for 2D RGB images, is able to improve segmentation quality for 3D multi-modal medical images.
    We developed an implementation of the algorithm which works for any number of spatial dimensions, input/output image channels, and reference image channels. 
    As far as we know this is the first publicly available implementation of this sort.
    We tested the algorithm with two distinct 3D medical imaging datasets, we concluded that the performance differences observed were not statistically significant.
    Finally, in the discussion section of the paper, we go into the reasons as to why this technique transfers poorly from natural images to medical images.  
\end{abstract}

\section{Introduction}

Using a fully-connected \gls{crf} \cite{FullyConnectedCRF} in conjunction with a \gls{fcnn} \cite{long2015fully} is the state-of-the art type of approach for semantic segmentation of two-dimensional natural images \cite{DeepLab}.

The core idea behind this approach is that the \gls{fcnn} will serve as a feature extractor that produces a coarse segmentation which is later refined by the \gls{crf}.
The \gls{crf} takes as input the segmentation produced by the network as well as the original input image.
Unlike a convolution layer which employs local filters, the \gls{crf} looks at every possible pair of pixels in the image, also known as a clique.
The \gls{crf} is a graphical model where every clique is defined not only by the spatial distance between pixels but also by their distance in colour space.
This allows the \gls{crf} to produce a segmentation with much sharper edges when compared with only using a \gls{fcnn}.

This means that the receptive field of a \gls{crf} is the entire image.

Graphical models such as the fully-connected \gls{crf} have also been extensevely used in 3D medical imaging segmentation with good resuts \cite{medseg_example_1, medseg_example_2, medseg_example_3, medseg_example_4, medseg_example_5}.
One of the issues with applying a fully-connected \gls{crf} to 3D images is the fact that the third dimension introduces exponentially more pairs of hyper-voxels in the graph, also known as cliques. For example, if a 2D square image with $width=N$ will have $N^2$ cliques in the graph, a 3D image with the same width will have $N^3$ cliques in the graph.
This makes these models more expensive and likely justifies why they are not as widespread in 3D medical images as in 2D images.

Another issue with using a \gls{crf} to improve the quality of a segmentation is that the \gls{crf} has to be trained separately after the base classifier has been trained. 
Because of this, in \cite{CRFasRNN} the authors propose writing the \gls{crf} mean-field approximation described in \cite{FullyConnectedCRF} as an \gls{rnn} which can be placed on top of \gls{cnn} and train the whole system end-to-end.

As far as we know, this system has not been tested for three-dimensional medical images.
As a result, we set out to test if this algorithm could be successfully applied in this domain.

\section{Methods}

\subsection{Theoretical background.}
In this section we present a summary of theoretical background of a fully-connected \gls{crf}\cite{FullyConnectedCRF}.

Consider an n-dimensional image with $N$ hyper-voxels (pixels, voxels, 4D-voxels, etc\dots) on which we wish to perform semantic segmentation, \textit{i.e.} to assign a label to every hyper-voxels.
We define $X_j$ and $I_j$ to be the label and colour value of hyper-voxel $j$, respectively.

Consider a random field $\textbf{X}$ defined over a set of variables $\{X_1, X_2, \dots, X_N\}$ each taking a value from a set of labels $\mathcal{L} = \{ l_1, l_2, \dots, l_k\}$.
Consider another random field $\textbf{I}$ defined over the variables $\{I_1, I_2, \dots, I_N\}$ where the domain of each variable is the possible colour values of a hyper-voxel in the image.

A \acrlong{crf} $(\textbf{I}, \textbf{X})$ is characterized by a Gibbs distribution:
\begin{equation}
    P(\textbf{X} | \textbf{I}) = \frac{1}{Z(\textbf{I})}\exp{ \left ( -\sum_{c \in \mathcal{C_G} }{ \phi_c(\textbf{X}_c| \textbf{I})} \right )},
\end{equation}
where $\mathcal{G}$ is a graph on $\textbf{X}$ and each clique $c$ in the set of cliques $\mathcal{C_G}$ induces a potential $\phi_c$. 
The Gibbs energy of labelling $\textbf{x} \in \mathcal{L}^N$ is $E(\textbf{x}|\textbf{I}) = \sum_{c \in \mathcal{C_G} }{ \phi_c(\textbf{X}_c| \textbf{I})} $ and the \gls{map} labelling of the random field is $\textbf{x}^\ast = \text{arg max}_{\textbf{x} \in \mathcal{L}^N} P(\textbf{X} | \textbf{I})$.
$Z(\textbf{I})$ is a normalization constant that ensures $P(\textbf{X} | \textbf{I})$ is a valid probability distribution.
For notational convenience the conditioning will be omitted from now on, we define $\psi_c(\textbf{x}_c)$ to denote $\phi_c(\textbf{x}_c| \textbf{I})$.

The Gibbs energy of the fully-connected pairwise \gls{crf} is the set of all unary and pairwise potentials:
\begin{equation}
    E(x) = \sum_{i}{\psi_u(x_i)} + \sum_{i<j}{\psi_p(x_i, x_j)},
\end{equation}
where $i$ and $j$ range from 1 to $N$.
The unary potential $\psi_u(x_i)$ is computed independently for each hyper-voxel by a classifier, \textit{i.e.} the choice of label for one hyper-voxel does have a direct impact on the labels of other hyper-voxels.

The pairwise potentials are given by:
 
\begin{equation}
    \psi_p(x_i, x_j) = \mu(x_i, x_j) \underbrace{\sum_{m=1}^{K}{w^{(m)}k^{(m)}(\textbf{f}_i, \textbf{f}_j)}}_{k(\textbf{f}_i, \textbf{f}_j)},
\end{equation}
where $k^{(m)}$ is a Gaussian kernel applied to arbitrary feature vectors $\textbf{f}_i$ and $\textbf{f}_j$, $w^{(m)}$ is linear combination of trainable weights and $mu$ is a compatibility function between labels.

The feature vectors $\textbf{f}_i$ and $\textbf{f}_j$ can be constructed from any feature space regarding the image. 
However, in this setting, they are chosen to take into account positions $p_i$ and $p_j$, and the colour values $I_i$ and $I_j$ of the hyper-voxels in the image:

\begin{equation}
    k(\textbf{f}_i, \textbf{f}_j) = w^{(1)}\underbrace{\exp{\left ( -\frac{|p_i-p_j|^2}{2\theta_\alpha^2} - \frac{|I_i-I_j|^2}{2\theta_\beta^2} \right )}}_{\text{appearance kernel}} + w^{(2)}\underbrace{\exp{\left ( - \frac{|p_i-p_j|^2}{2\theta_\gamma^2}\right )}}_{\text{smoothness kernel}}.
\end{equation}

The parameters $\theta_\alpha$, $\theta_\beta$ and $\theta_\gamma$ are hyper-parameters that control the importance of the hyper-voxel difference in a specific feature space.
This choice of $k(\textbf{f}_i, \textbf{f}_j)$ includes both an appearance kernel (aka bilateral kernel), which penalizes different labels for hyper-voxels that are close in space and color value, and a smoothness kernel (aka Gaussian kernel) which penalizes different labels for hyper-voxels close only in space.

In our case, the compatibility function, $\mu$, is a $k$ by $k$ matrix learnt from the data.
It has zeros along its diagonal and trainable weights elsewhere in order for the model to be able to penalize different pairs of labels differently.
For instance, in brain tumour segmentation we might want to penalize assigning the background class to the tumour's core more than assigning the oedema class to the tumour's core.

Since the direct computation of $P(\textbf{X})$ is intractable we use the mean field approximation to compute the distribution $Q(\textbf{X})$ that minimizes the KL-divergence $\textbf{D}(Q||P)$, where $Q$ can be written as a product of independent marginals, $Q(\textbf{X})=\prod_i Q_i(X_i)$.

Minimizing the KL-divergence yields the following iterative update equation:

\begin{equation}
    Q_i(x_i=l) = \frac{1}{Z_i}\exp{\left ( -\psi_u(x_i) - \sum_{l' \in \mathcal{L}}\mu(l, l') \sum_{m=1}^{K}w^{(m)} \sum_{i \neq j} k^{(m)}(\textbf{f}_i, \textbf{f}_j) Q_j(l')\right )},
\end{equation}
which leads to the inference algorithm detailed in Algorithm~\ref{alg:mean_field}.

\begin{algorithm}
\SetAlgoLined
 $Q_i(x_i) = \frac{1}{Z_i}\exp\{-\phi_u(x_i)\}$;  \hfill Initialize $Q$ 
 
 \While{not converged}{
 
 $\tilde{Q}^{(m)}_i(l) \leftarrow \sum_{i \neq j}{k^{(m)}(\textbf{f}_i, \textbf{f}_j)Q_j(l)}$ for all $m$; \hfill Message passing
 
 $\hat{Q}_i(x_i) \leftarrow \sum_{l \in \mathcal{L}}{\mu^{(m)}(x_i, l)\sum_{m}{w^{(m)}}\tilde{Q}_i^{(m)}(l)}$; \hfill Compatibility transform
 
 $Q_i(x_i) \leftarrow \exp\{-\psi(xi) - \hat{Q}_i(x_i)$\}; \hfill Local update
 
 normalize $\hat{Q}_i(x_i)$
 }
 \caption{\gls{crf} mean field approximation.}
 \label{alg:mean_field}
\end{algorithm}

The only step in Algorithm~\ref{alg:mean_field} that is not straightforward is the message passing step from every $X_i$ to $X_j$. 
For our choice of kernels, this step involves applying a bilateral and a Gaussian filter to $Q$.
A brute force implementation has a time complexity of $\mathcal{O}(N^2)$, therefore, we use the permutohedral lattice to approximate high-dimensional filtering \cite{adams2010fast} which has linear time complexity with $N$ (even though it has quadratic time complexity with the number of dimensions of the position vectors).

The key insight of the \gls{crf} as \gls{rnn} paper \cite{CRFasRNN} is that this inference algorithm can be written as a sequence of steps which can propagate gradient backwards like a \gls{rnn}.
This can be easily implemented in an existing deep learning framework.
The authors called this new layer a \gls{crf} as \gls{rnn} layer which can be placed on top of existing \gls{cnn} architectures to improve the quality of semantic segmentation.
The main advantage of this layer is the ability to train a model which includes a \gls{crf} end-to-end with gradient descent methods.

\subsection{Implementation}

The previously proposed system was designed and implemented for 2D RGB images.
In our work, we adapted the algorithm to work with n-dimensional images and with any number of channels.

From a conceptual stand-point extending this algorithm to the general case is straightforward.
However, the implementation details are a bit more complicated and hence are the core contribution of this work.
Most of the steps in the inference algorithm can be easily written using existing operations in popular deep learning frameworks.
Unfortunately, the message passing step which involves high-dimensional filtering cannot be easily implemented using existing operations.

Both public implementations of the fully-connected \gls{crf} and \gls{crf} as \gls{rnn} algorithms are based on the permutohedral lattice algorithm \cite{adams2010fast} and the code provided in that paper.
The permutohedral lattice is a fast approximation of high dimensional filters which can be used for Gaussian and bilateral filtering.
The available implementation of the permutohedral lattice was designed for 2D RGB images and only used CPU kernels \footnote{The authors of the permutohedral lattice paper also provided a GPU implementation which proved to be slower than the CPU version due to bugs in the code.}.

The brunt of this work was re-implementing the permutohedral lattice so that:
\begin{itemize}
    \item The implementation supported any number of spatial dimensions, input channels/output channels (class labels from the perspective of the \gls{crf}) and reference image channels. 
    \item The implementation contained not only a CPU C++ kernel but also as a C++/CUDA kernel for fast computation in GPU.
    \item The implementation included a TensorFlow Op wrapper so that it could be easily used in Python and incorporated in any TensorFlow graph.
\end{itemize}

Our code for the permutohedral lattice (both CPU and GPU) implemented as a TensorFlow operation is available at \url{https://github.com/MiguelMonteiro/permutohedral_lattice}.
The code for the \gls{crf} as \gls{rnn} algorithm also implemented in TensorFlow and which uses the aforementioned permutohedral lattice is available at \url{https://github.com/MiguelMonteiro/CRFasRNNLayer}.

\section{Experiments}

To test whether using the \gls{crf} as \gls{rnn} layer on top of a \gls{cnn} improved the segmentation quality for 3D medical images, we conducted two experiments.

The aim of these experiments was not to achieve the best possible performance on these tasks, but to compare the performance difference between using and not using the proposed algorithm.

The underlying network architecture used for segmentation was the V-Net architecture \cite{VNet}.

\subsection{PROMISE 2012}

\begin{figure}[ht]
\centering
\begin{subfigure}{.25\textwidth}
  \centering
  \includegraphics[width=\linewidth]{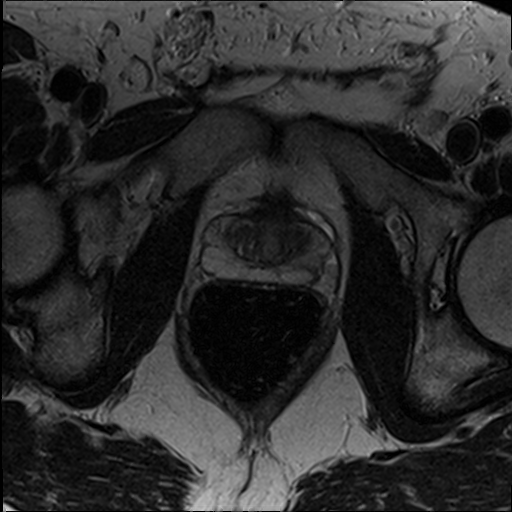}
  \caption{Image.}
\end{subfigure}%
\begin{subfigure}{.25\textwidth}
  \centering
  \includegraphics[width=\linewidth]{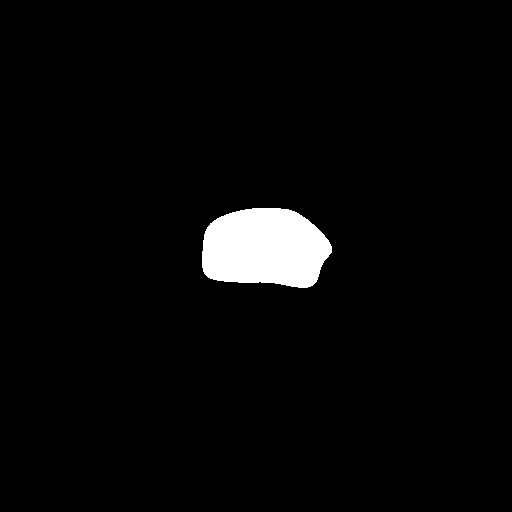}
  \caption{Labels.}
\end{subfigure}
\caption{PROMISE 2012 example slice.}
\label{fig:promise_2012}
\end{figure}

The PROMISE 2012 dataset \cite{PROMISE_2012} is a set of 50 three-dimensional \gls{mri} prostate images and the respective expert segmentations of the prostate. 
The images have different resolutions and a different number of slices, regardless, they all have one channel.
In our experiment, we re-sampled the images so that they all had resolution $1\times1\times2$ millimetres.
This resulted in images of size $[x, y, z, c] = [200, 200, 63, 1]$ voxels, where $c$ denotes the channel dimension.
An example slice from the PROMISE 2012 dataset along with the respective expert segmentation is shown in Figure~\ref{fig:promise_2012}.

Since this is a binary segmentation problem the performance metric used is simply the \gls{dc}.

Given that there were only 50 labelled images we used 5-fold cross-validation and ran each fold for 200 epochs (8000 steps). The results for this experiment with and without the \gls{crf} as \gls{rnn} layer are presented in Table~\ref{tab:promise_2012}.

\begin{table}[!ht]
\centering
\caption{PROMISE 2012 results}
\label{tab:promise_2012}
\begin{tabular}{@{}ccc@{}}
\toprule
            & Dice Coefficient   \\ \midrule
Without CRF & $0.767 \pm 0.109$  \\
With CRF    & $0.780 \pm 0.110$  \\ \bottomrule
\end{tabular}
\end{table}

\subsection{BraTS 2015 High-Grade Gliomas}

The \gls{brats_2015} \cite{BRATS_2015} training dataset for \gls{hgg} is composed of 220 multimodal \gls{mri} images of brain tumors.
All of the images have the same resolution $1\times 1 \times 1$ millimeters and the same size ($[x, y, z] = [240, 240, 155]$ voxels). 
For each case there exist 4 different images (T1, T1c, T2 and Flair).
This results in images of size $[x, y, z, c] = [240, 240, 155, 4]$.

\begin{figure}[ht]
\centering
\begin{subfigure}{.2\textwidth}
  \centering
  \includegraphics[width=\linewidth]{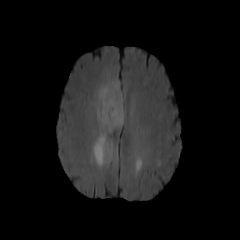}
  \caption{Flair.}
\end{subfigure}%
\begin{subfigure}{.2\textwidth}
  \centering
  \includegraphics[width=\linewidth]{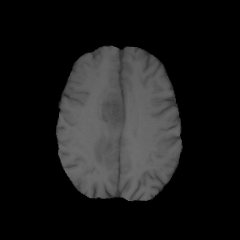}
  \caption{T1.}
\end{subfigure}%
\begin{subfigure}{.2\textwidth}
  \centering
  \includegraphics[width=\linewidth]{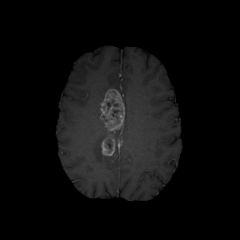}
  \caption{T1C.}
\end{subfigure}%
\begin{subfigure}{.2\textwidth}
  \centering
  \includegraphics[width=\linewidth]{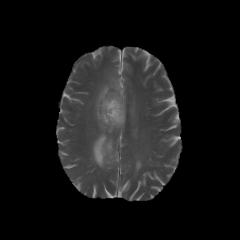}
  \caption{T2.}
\end{subfigure}%
\begin{subfigure}{.2\textwidth}
  \centering
  \includegraphics[width=\linewidth]{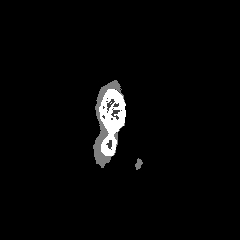}
  \caption{Labels.}
\end{subfigure}
\caption{BraTS 2015 HGG example slice.}
\label{fig:brats_2015}
\end{figure}

The expert segmentation has 5 distinct labels: background, oedema, enhancing tumour core, non-enhancing tumour core and necrotic tumour core.
However, the main performance metrics for this task are the whole tumour \gls{dc} (includes everything except the background) and the core tumour \gls{dc} (only includes the enhancing, non-enhancing and necrotic cores).
An example slice from the BraTS 2015 dataset along with the respective expert segmentation is shown in Figure~\ref{fig:brats_2015}.

For our experiments, we split the data-set into training and holdout set (85\%/15\%) which meant that 33 cases were used to measure performance on.
The network was trained for 100 epochs (18700 steps).
The results for this experiment with and without the \gls{crf} as \gls{rnn} layer are presented in Table~\ref{tab:brats_2015}.

\begin{table}[!h]
\centering
\caption{\gls{brats_2015} \gls{hgg} Results}
\label{tab:brats_2015}
\begin{tabular}{@{}ccc@{}}
\toprule
            & Whole Tumor Dice Coefficient        & Core Tumor Dice Coefficient         \\ \midrule
Without CRF & $0.735 \pm 0.105$   & $0.488 \pm 0.244$                                   \\
With CRF    & $0.738 \pm 0.105$   & $0.482 \pm 0.257$                                   \\ \bottomrule
\end{tabular}
\end{table}

\section{Discussion}

Looking at close means and large standard deviations presented in Table~\ref{tab:promise_2012} we can see that it is unlikely that a statistical test will reveal that the small performance increase in using the \gls{crf} is statistically significant. 
In fact, a paired t-test reveals exactly this: the performance difference between using and not using the \gls{crf} as \gls{rnn} is not statistically significant. 
The same is true for the results of the BraTS 2015 experiment, presented in Table~\ref{tab:brats_2015}.

Hence, we conclude that using the \gls{crf} as \gls{rnn} layer on top of a \gls{cnn} does not improve the segmentation quality.
The fact that this algorithm seemingly works for 2D RBG images \cite{CRFasRNN} but not for 3D \gls{mri} medical images can be due to a number of factors. 
Here we explore some of those factors.

Natural images tend to have much higher contrast and much sharper edges than \gls{mri} images.
The edges between objects in natural images tend to be much more well defined (e.g. A building against a blue sky) than in \gls{mri} images (e.g. the oedema in a brain \gls{mri} is a slightly different shade of grey than the healthy region surrounding it).
Since \gls{mri} images have much less contrast and tend to have blurry edges, the object of interest often fuses with the background slowly and seamlessly 
Trained radiologists can use their knowledge of human anatomy and pathology in conjunction with the observed image to infer where the object of interest starts and ends.
In contrast, the \gls{crf} only has access to differences between hyper-voxels, and these differences are zero or close to zero in low contrast, blurry edge environments.
This means that there is much more sensitivity to the parameters $\theta_\alpha$, $\theta_\beta$ and $\theta_\gamma$. Setting these parameters becomes very difficult and when taking into account inter-image variability observed during training, there simply isn't a set of $\theta$ parameters that works for all (or most of) the images. 
    
Despite having many voxels, 3D medical images have much lower resolution then natural images which compounds the problem of having blurry edges.

The regions of interest to be segmented in medical images tend to be ``local''. In our experiment, the prostate and brain tumours fit inside the receptive field of the neural network.
This may not be the case in natural images where we might, for example, want to segment multiple birds out of the sky. 
For this reason, it is possible that the \gls{fcnn} was already able to capture all of the relevant spatial and colour relations in the image and hence the \gls{crf} has no room to improve.

\begin{figure}
\centering
\begin{subfigure}{.25\textwidth}
  \centering
  \includegraphics[width=\linewidth]{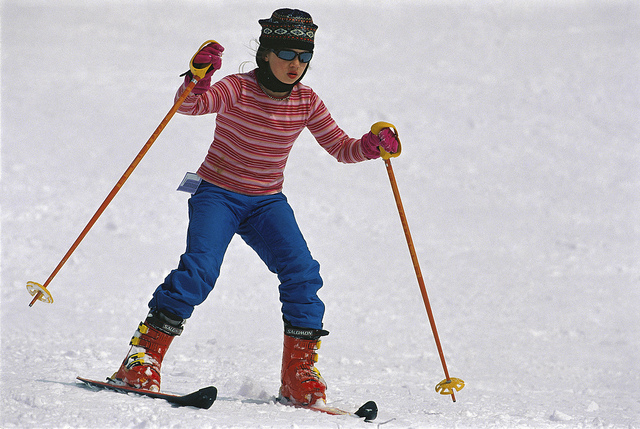}
  \caption{\centering Image. \protect\linebreak}
\end{subfigure}%
\begin{subfigure}{.25\textwidth}
  \centering
  \includegraphics[width=\linewidth]{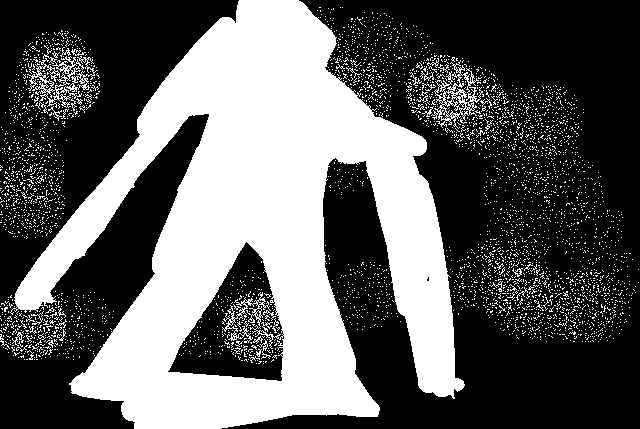}
  \caption{\centering Distorted segmentation.}
\end{subfigure}%
\begin{subfigure}{.25\textwidth}
  \centering
  \includegraphics[width=\linewidth]{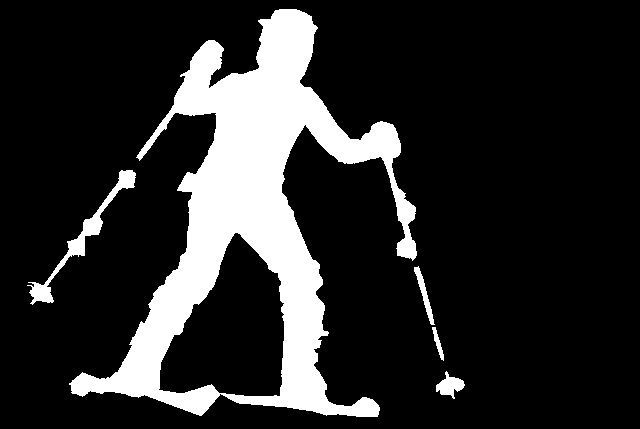}
  \caption{\centering Correct segmentation.}
\end{subfigure}%
\begin{subfigure}{.25\textwidth}
  \centering
  \includegraphics[width=\linewidth]{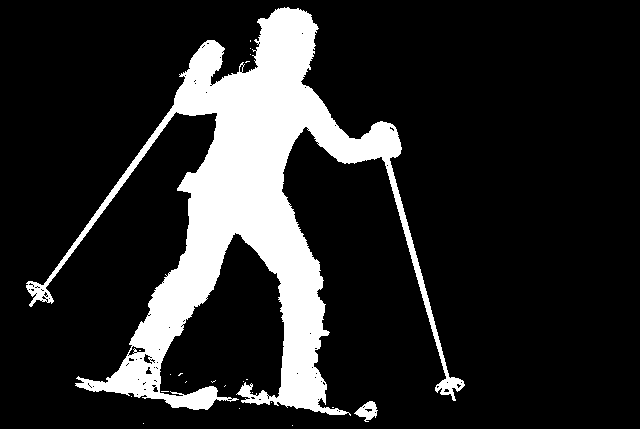}
  \caption{\centering Predicted segmentation.}
\end{subfigure}%
\caption{2D RGB experiment.}
\label{fig:rgb_experiment}
\end{figure}

One thing the reader might wondering at this point is whether there was a bug with our implementation. 
To test this proposition we tested our implementation of the algorithm for 2D RGB images. 
We took an image and its respective segmentation, we distorted the segmentation to simulate the output of a "bad" \gls{cnn}.
After, we overfitted the \gls{crf} as \gls{rnn} layer by giving it as input the distorted segmentation
and minimizing the cross-entropy between the output of the \gls{crf} as \gls{rnn} layer and the correct labels.
The results of this experiment are shown in Figure~\ref{fig:rgb_experiment}. 
As we can from Figure~\ref{fig:rgb_experiment} the predicted segmentation that is accomplished just by training two scalar weights on filter outputs and running the recurrent inference algorithm gives better results than even the expert provided segmentation (as seen from the skier's baton).
This indicates that both the filters and inference algorithm are implemented correctly.

We also performed the same experiment for 3D medical images from the PROMISE 2012 and BraTS 2015 dataset and observed that the \gls{crf} as \gls{rnn} layer was able to remove the greater part of the noise. However, the algorithm was not able to overfit the visual features nearly as well as in the 2D RGB case.

\section{Conclusion}

In this paper we applied the \gls{crf} as \gls{rnn} layer for semantic segmentation to 3D medical imaging segmentation. 
As far as we know we provide the first publicly available version of this algorithm that works for any number of spatial dimensions, input/output channels and reference channels.
We tested the \gls{crf} as \gls{rnn} layer on top a \gls{fcnn} (with a V-Net architecture) on two distinct medical imaging datasets.
We concluded that the performance differences observed were not statically significant and we provide a discussion as to why this technique does not transfer well from 2D RGB images to 3D multi-modal medical images.

\bibliographystyle{plain}
\bibliography{bibliography.bib}

\end{document}